\def\year{2022}\relax
\documentclass[letterpaper]{article} 
\usepackage{aaai22}  
\usepackage{times}  
\usepackage{helvet}  
\usepackage{courier}  
\usepackage[hyphens]{url}  
\usepackage{graphicx} 
\usepackage{natbib}  
\usepackage{caption} 
\DeclareCaptionStyle{ruled}{labelfont=normalfont,labelsep=colon,strut=off} 
\usepackage{algorithm}
\usepackage{algorithmic}

\usepackage{amsmath}
\usepackage{amsfonts}
\usepackage{bm}

\usepackage{booktabs}
\usepackage{threeparttable}

\usepackage[switch]{lineno}
\usepackage{multirow}

\usepackage{lineno}

\usepackage{color}

\usepackage{newfloat}
\usepackage{listings}
\floatstyle{ruled}
\newfloat{listing}{tb}{lst}{}
\floatname{listing}{Listing}
\title{Shadow Generation for Composite Image in Real-World Scenes}

\author{Yan Hong$^{1}$, Li Niu$^{1}$\thanks{Corresponding author.}, Jianfu Zhang$^{2}$}
\affiliations{
   $^{1}$ MoE Key Lab of Artificial Intelligence, Department of Computer Science and Engineering \\
Shanghai Jiao Tong University, Shanghai, China \\
$^{2}$ Tensor Learning Team, RIKEN AIP, Tokyo, Japan \\
yanhong.sjtu@gmail.com, ustcnewly@sjtu.edu.cn, jianfu.zhang@riken.jp.
}

\usepackage{bibentry}

\begin{document}

\nolinenumbers
\maketitle


\begin{abstract}
Image composition targets at inserting a foreground object into a background image. Most previous image composition methods focus on adjusting the foreground to make it compatible with background while ignoring the shadow effect of foreground on the background. 
In this work, we focus on generating plausible shadow for the foreground object in the composite image. First, we contribute a real-world shadow generation dataset DESOBA by generating synthetic composite images based on paired real images and deshadowed images.  Then, we propose a novel shadow generation network SGRNet, which consists of a shadow mask prediction stage and a shadow filling stage. In the shadow mask prediction stage, foreground and background information are thoroughly interacted to generate foreground shadow mask. In the shadow filling stage, shadow parameters are predicted to fill the shadow area. Extensive experiments on our DESOBA dataset and real composite images demonstrate the effectiveness of our proposed method. 
\end{abstract}

\section{Introduction}
 \label{sec:introduction}
 
Image composition \cite{niu2021making} targets at copying a foreground object from one image and pasting it on another background image to produce a composite image. 
In recent years, image composition has drawn increasing attention from a wide range of applications in the fields of medical science, education, and entertainment~\cite{arief2012realtime, zhang2019shadowgan,liu2020arshadowgan}. Some deep learning methods~\cite{lin2018st, azadi2020compositional, van2020investigating, compgan} have been developed to improve the realism of composite image in terms of color consistency, relative scaling, spatial layout, occlusion, and viewpoint transformation. However, the above methods mainly focus on adjusting the foreground while neglecting the effect of inserted foreground on the background such as shadow or reflection. In this paper, we focus on dealing with the shadow inconsistency between the foreground object and the background, that is, generating shadow for the foreground object according to background information, to make the composite image more realistic.

To accomplish this image-to-image translation task, deep learning techniques generally require adequate paired training data, \emph{i.e.}, a composite image without foreground shadow and a target image with foreground shadow.
However,  it is extremely difficult to obtain such paired data in the real world. Therefore, previous works~\cite{zhang2019shadowgan,liu2020arshadowgan} insert a virtual 3D object into 3D scene and generate shadow for this object using rendering techniques. In this way, a rendered dataset with paired data can be constructed. However, there exists large domain gap between rendered images and real-world images, which results in the inapplicability of rendered dataset to real-world image composition problem. 

\begin{figure}
\begin{center}
\includegraphics[scale=0.36]{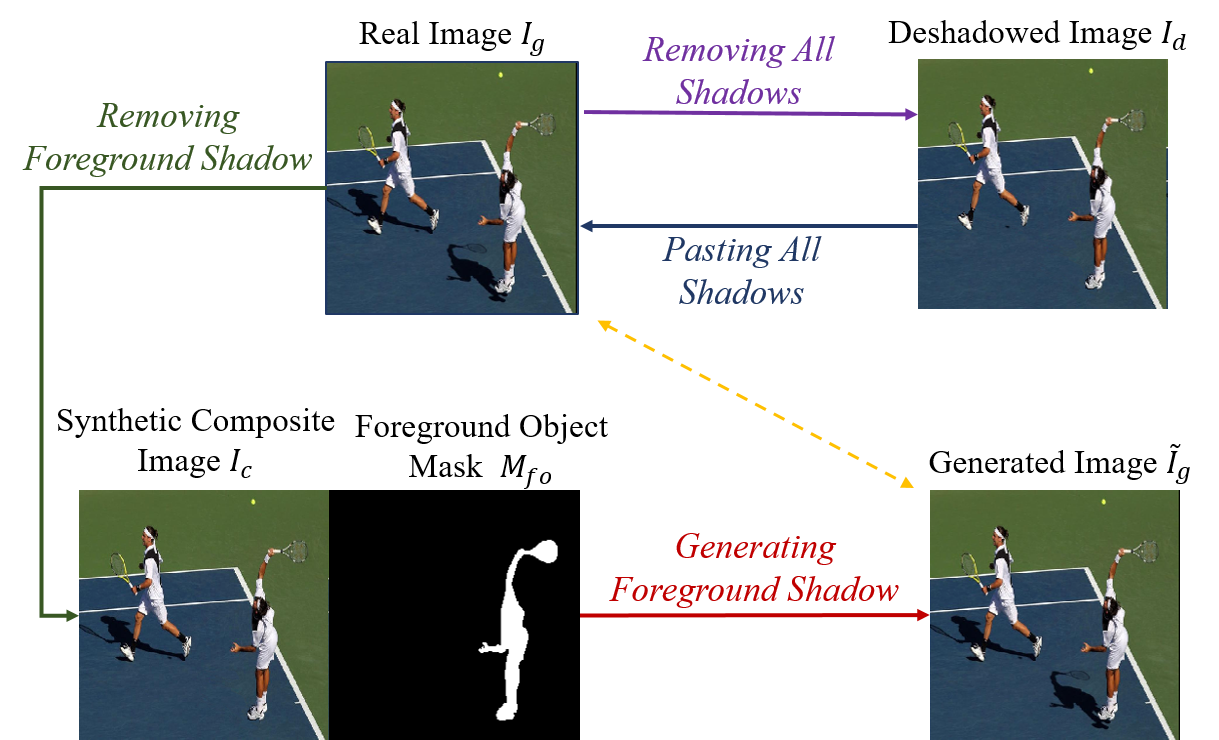}
\end{center}
\caption{1) The green arrows illustrate the process of acquiring paired data.  We select a foreground object in the ground-truth target image $\mathbf{I}_g$, and replace its shadow area with the counterpart in deshadowed image $\mathbf{I}_{d}$ to synthesize a composite image $\mathbf{I}_c$. 
2) The red arrow illustrates our shadow generation task. Given $\mathbf{I}_c$ and its foreground mask $\bm{M}_{fo}$, we aim to generate the target image $\tilde{\mathbf{I}}_g$ with foreground shadow.}
\label{fig:task_intro} 
\end{figure}

Therefore, we tend to build our own real-world shadow generation dataset by synthesizing composite image from a ground-truth target image with object-shadow pairs. We build our dataset on the basis of Shadow-OBject Association (SOBA) dataset~\cite{wang2020instance}, which collects real-world images in complex scenes and provides annotated masks for object-shadow pairs. SOBA contains 3,623 pairs of shadow-object associations over 1000 images. Based on SOBA dataset, we remove all the shadows to construct our DEshadowed SOBA (DESOBA) dataset, which can be used for shadow generation task as well as other relevant vision applications. At the start, we tried to remove the shadows with the state-of-the-art deshadow methods~\cite{zhang2020ris, le2020shadow, cun2020towards}. However, their performance is far from satisfactory due to complex scenes. Thus,  with shadow images and shadow masks from SOBA datasets, we employ professional photo editors to manually remove the shadows in each image to obtain deshadowed images. We carefully check each deshadowed image to ensure that the background texture is maintained to the utmost, the transition over shadow boundary is smooth, and the original shadowed area cannot be identified. Although the deshadowed images may not be perfectly accurate, we show that the synthetic dataset is still useful for method comparison and real-world image composition. 
One example of ground-truth target image $\mathbf{I}_g$ and its deshadowed version $\mathbf{I}_d$ is shown in Figure~\ref{fig:task_intro}. To obtain paired training data for shadow generation task, we choose a foreground object with associated shadow in the ground-truth target image $\mathbf{I}_g$ and replace its shadow area with the counterpart in the deshadowed image $\mathbf{I}_d$, yielding the synthetic composite image $\mathbf{I}_c$. In this way, pairs of synthetic composite image $\mathbf{I}_c$ and ground-truth target image $\mathbf{I}_g$ can be obtained. 

With paired training data available, the shadow generation task can be defined as follows. \emph{Given an input composite image $\mathbf{I}_c$ and the foreground object mask $\mathbf{M}_{fo}$, the goal is to generate realistic shadow for the foreground object, resulting in the target image $\tilde{\mathbf{I}}_g$ which should be close to the ground-truth $\mathbf{I}_g$} (see Figure~\ref{fig:task_intro}).
For ease of description, we use foreground (\emph{resp.}, background) shadow to indicate the shadow of foreground (\emph{resp.}, background) object. 
Existing image-to-image translation methods~\cite{pix2pix2017, zhu2017unpaired, huang2018multimodal, lin2018conditional} can be used for shadow generation, but they cannot achieve plausible shadows without considering illumination condition or shadow property. ShadowGAN~\cite{zhang2019shadowgan} was designed to generate shadows for virtual objects by combining a global discriminator and a local discriminator. ARShadowGAN~\cite{liu2020arshadowgan} searched clues from background using attention mechanism to assist in shadow generation. However, the abovementioned methods did not model the thorough foreground-background interaction and did not leverage typical illumination model, which motivates us to propose a novel Shadow Generation in the Real-world Network (SGRNet) to generate shadows for the foreground objects in complex scenes.

As illustrated in Figure~\ref{fig:framework}, \emph{SGRNet consists of a shadow mask prediction stage and a shadow filling stage.} Such two-stage approach has not been explored in shadow generation task before. In the shadow mask prediction stage, provided with a synthetic composite image $\mathbf{I}_c$ and foreground object mask $\mathbf{M}_{fo}$, we design a foreground encoder to extract the required information of the foreground object and a background encoder to infer illumination information from background. To achieve thorough foreground and background information interaction, a cross-attention integration layer is employed to help generate shadow mask for the foreground object. The shadow filling stage is designed based on illumination model~\cite{le2019shadow}, which first predicts the shadow property and then edits the shadow area. Besides, we design a conditional discriminator to distinguish real object-shadow-image triplets from fake triplets, which can push the generator to produce realistic foreground shadow. To verify the effectiveness of our proposed SGRNet, we conduct experiments on DESOBA dataset and real composite images. Our dataset and code are available at https://github.com/bcmi/Object-Shadow-Generation-Dataset-DESOBA.

Our main contributions are summarized as follows: 1) we contribute the first real-world shadow generation dataset DESOBA using a novel data acquisition approach; 2) we design a novel two-stage network SGRNet to generate shadow for the foreground object in composite image; 3) extensive experiments demonstrate the effectiveness of our way to construct dataset and the superiority of our network.


\begin{figure*}
\begin{center}
\includegraphics[scale=0.58]{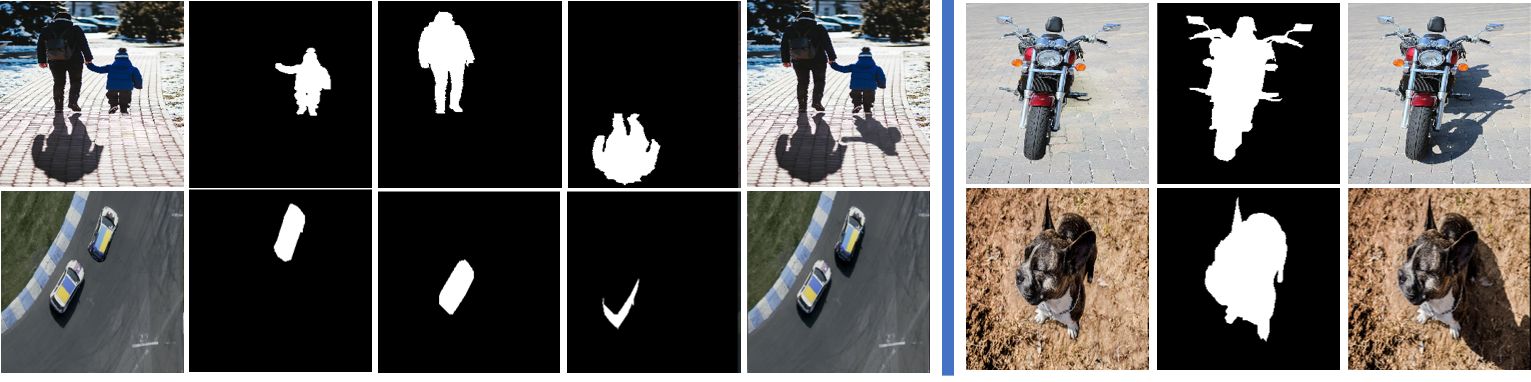}
\end{center}
\caption{Some examples from our DESOBA dataset. BOS test image pairs with Background Object-Shadow (BOS) pairs are shown in the left subfigure, from left to right: synthetic composite image, foreground object mask, background object mask, background shadow mask, and ground-truth target image. BOS-free test image pairs are shown in the right subfigure, from left to right: synthetic composite image, foreground object mask, and ground-truth target image.}
\label{fig:dataset_samples} 
\end{figure*}

\section{Related Work}
\label{sec:related}
\subsection{Image Composition}
Image composition \cite{niu2021making} targets at pasting a foreground object on another background image to produce a composite image~\cite{lin2018st,wu2019gp, zhan2019adaptive, zhan2020towards, liu2020arshadowgan}. Many issues would significantly degrade the quality of composite images, such as unreasonable location of foreground or inconsistent color/illumination between foreground and background. Previous works attempted to solve one or multiple issues. For example, image blending methods~\cite{perez2003poisson,wu2019gp,zhang2020deep,zhang2021deep} were developed to blend foreground and background seamlessly. Image harmonization methods~\cite{tsai2017deep, xiaodong2019improving, cong2020dovenet,cong2021bargainnet} were proposed to address the color/illumination discrepancy between foreground and background. 
Some other approaches \cite{chen2019toward,weng2020misc,zhan2019spatial} aimed to cope with the inconsistency of geometry, color, and boundary simultaneously.
However, these methods did not consider the shadow effect of inserted foreground on background image, which is the focus of this paper.

\subsection{Shadow Generation}
Prior works on shadow generation can be divided into two groups: rendering based methods and image-to-image translation methods. 

\noindent\textbf{Shadow Generation via Rendering:} This group of methods require explicit knowledge of illumination, reflectance, material properties, and scene geometry to generate shadow for inserted virtual object using rendering techniques. However, such knowledge is usually unavailable in the real-world applications. Some methods~\cite{karsch2014automatic, 2014Exposing, liu2017static} relied on user interaction to acquire illumination condition and scene geometry, which is time-consuming and labor-intensive. Without user interaction, some methods~\cite{liao2019illumination,gardner2019deep,zhang2019all, arief2012realtime} attempted to recover explicit illumination condition and scene geometry based on a single image, but this estimation task is quite tough and inaccurate estimation may lead to terrible results~\cite{zhang2019shadowgan}.

\noindent\textbf{Shadow Generation via Image-to-image Translation:} This group of methods learn a mapping from the input image without foreground shadow to the output with foreground shadow, without requiring explicit knowledge of illumination, reflectance, material properties, and scene geometry. Most methods within this group have encoder-decoder network structures. For example, the shadow removal method Mask-ShadowGAN~\cite{hu2019mask} could be adapted to shadow generation, but the cyclic generation procedure failed to generate shadows in complex scenes. ShadowGAN~\cite{zhang2019shadowgan} combined a global conditional discriminator and a local conditional discriminator to generate shadow for inserted 3D foreground objects without exploiting background illumination information. In~\cite{zhan2020adversarial}, an adversarial image composition network was proposed for harmonization and shadow generation simultaneously, but it calls for extra indoor illumination dataset~\cite{gardner2017learning, illuminationdata}. ARShadowGAN~\cite{liu2020arshadowgan} released Shadow-AR dataset and proposed an attention-guided network. Distinctive from the above works, our proposed SGRNet encourages thorough information interaction between foreground and background, and also leverages typical illumination model to guide network design.


\section{Dataset Construction} \label{sec:dataset_cons}
We follow the training/test split in SOBA dataset~\cite{wang2020instance}. SOBA has $840$ training images with $2,999$ object-shadow pairs and $160$ test images with  $624$ object-shadow pairs. We discard one complex training image whose shadow is hard to remove.  Since most images in SOBA are outdoor images, we focus on outdoor illumination in this work. For each image in the training set, to obtain more training image pairs, we use a subset of foreground objects with associated shadows each time. Specifically, given a real image $\mathbf{I}_g$ with $n$ object-shadow pairs $\{(\mathbf{O}_i, \mathbf{S}_i)|_{i=1}^n\}$ and its deshadowed version $\mathbf{I}_d$ without shadows  $\{\mathbf{S}_i|_{i=1}^n\}$, we randomly select a subset of foreground objects from $\mathbf{I}_g$ and replace their shadow areas with the counterparts in $\mathbf{I}_d$, leading to a synthetic composite image $\mathbf{I}_c$. In this way, based on the training set of SOBA, we can obtain abundant training image pairs of synthetic composite images and ground-truth target images. In Section~\ref{sec:method}, for ease of description, we treat a subset of foreground objects as one whole foreground object.

For the test set, we can get pairs of synthetic composite images and ground-truth target images in the same way. We focus on synthetic composite images with only one foreground object and ignore those with too small foreground shadow after the whole image is resized to $256\times 256$. Afterwards, we obtain $615$ test image pairs,
which are divided into two groups according to whether they have background object-shadow pairs. 
Specifically, \emph{we refer to the test image pairs with Background Object-Shadow (BOS) pairs as BOS test image pairs, and the remaining ones as BOS-free test image pairs.} Despite the absence of strong cues like background object-shadow pairs, the background in BOS-free images could also provide a set of illumination cues (\emph{e.g.}, shading, sky appearance variation)~\cite{lalonde2012estimating, zhang2019all}.
Some examples of BOS test image pairs and BOS-free test image pairs are shown in Figure~\ref{fig:dataset_samples}. 

\section{Our Method}  \label{sec:method}
\begin{figure*}
\begin{center}
\includegraphics[scale=0.41]{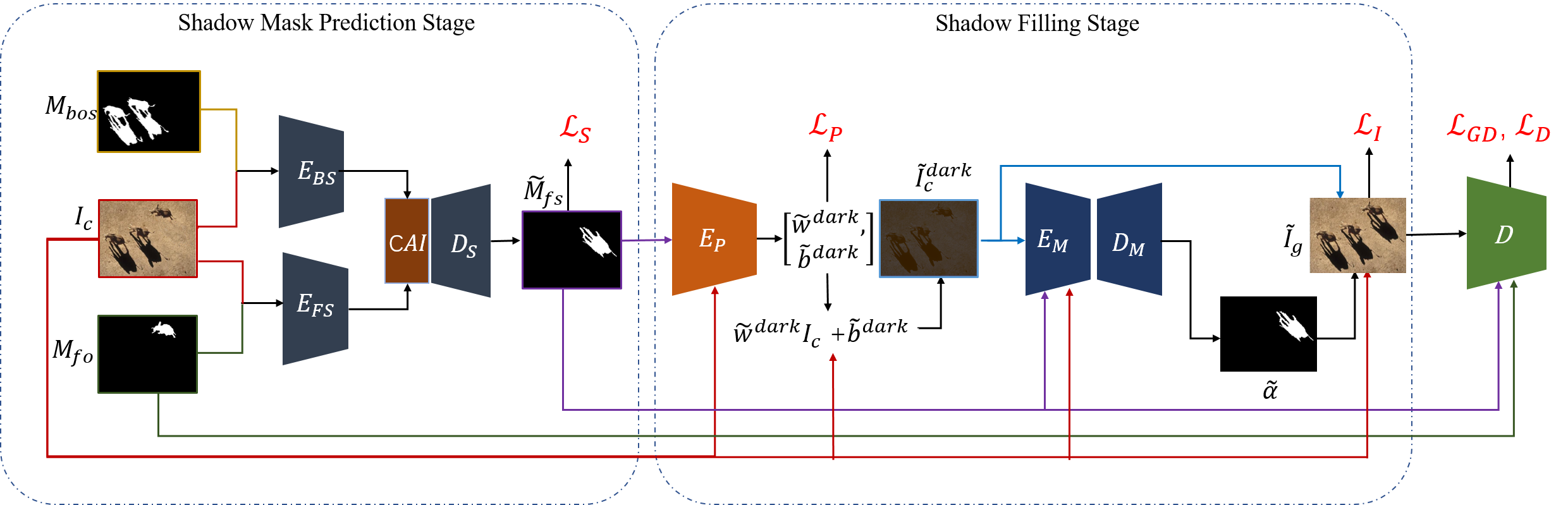}
\end{center}
\caption{The framework of our SGRNet which consists of a shadow mask prediction stage and a shadow filling stage. In the shadow mask prediction stage, shadow mask $\tilde{\bm{M}}_{fs}$ is generated by the shadow mask generator composed of foreground encoder $E_{FS}$, background encoder $E_{BS}$, Cross-Attention Integration (CAI) layer, and decoder $D_S$. In the shadow filling stage, shadow parameters $\{\tilde{\bm{w}}^{dark}, \tilde{\bm{b}}^{dark}\}$ are predicted by $E_{P}$ for producing darkened image $\tilde{\mathbf{I}}^{dark}_c$, and shadow matte predictor $G_M=\{E_M, D_M\}$ generates shadow matte $\tilde{\bm{\alpha}}$. The final target image $\tilde{\mathbf{I}}_g$ is obtained by blending $\tilde{\mathbf{I}}^{dark}_c$ and $\mathbf{I}_c$ using $\tilde{\bm{\alpha}}$. }
\label{fig:framework} 
\end{figure*}

Given a synthetic composite image $\mathbf{I}_c$ without foreground shadow and the foreground object mask ${\bm{M}}_{fo}$, our proposed Shadow Generation in the Real-world Network (SGRNet) targets at generating $\tilde{\mathbf{I}}_g$ with foreground shadow. Our SGRNet consists of two stages: a shadow mask prediction stage and a shadow filling stage (see Figure~\ref{fig:framework}). This two-stage approach enables the network to focus on one aspect (\emph{i.e.}, shadow shape or shadow intensity) in each stage, which has not been explored in previous shadow generation methods~\cite{zhang2019shadowgan,zhan2020adversarial,liu2020arshadowgan}. In the shadow mask prediction stage, a shadow mask generator $G_S$ with foreground branch and background branch is designed to generate shadow mask $\tilde{\bm{M}}_{fs}$. In the shadow filling stage, a shadow parameter predictor $E_{P}$ and a shadow matte generator $G_{M}$ are used to fill the shadow mask to produce the target image $\tilde{\mathbf{I}}_g$ with foreground shadow. 

\subsection{Shadow Mask Generator}
The shadow mask generator $G_S$ aims to predict the binary shadow mask $\tilde{\bm{M}}_{fs}$ of the foreground object. We adopt U-Net~\cite{ronneberger2015u} structure consisting of an encoder $E_S$ and a decoder $D_S$. 
To better extract foreground and background information,  we split $E_S$ into a foreground encoder $E_{FS}$ and a background encoder $E_{BS}$.
The foreground encoder $E_{FS}$ takes the concatenation of input composite image $\mathbf{I}_c$ and foreground object mask $\bm{M}_{fo}$ as input, producing the foreground feature map $\bm{X}_f = E_{FS}(\mathbf{I}_c, \bm{M}_{fo})$.
The background encoder $E_{BS}$ is expected to infer implicit illumination information from background. 
Considering that the background object-shadow pairs can provide strong illumination cues, we introduce background object-shadow mask $\bm{M}_{bos}$ enclosing all background object-shadow pairs.
The background encoder $E_{BS}$ takes the concatenation of $\mathbf{I}_c$ and $\bm{M}_{bos}$ as input, producing the background feature map $\bm{X}_b = E_{BS}(\mathbf{I}_c, \bm{M}_{bos})$. 

The illumination information in different image regions may vary due to complicated scene geometry and light sources, which greatly increases the difficulty of shadow mask generation~\cite{zhang2019shadowgan}. Thus, it is crucial to attend relevant illumination information to generate foreground shadow. Inspired by previous attention-based methods~\cite{zhang2019self,wang2018non,vaswani2017attention}, we use a Cross-Attention Integration (CAI) layer to help foreground feature map $\bm{X}_f$ attend relevant illumination information from background feature map $\bm{X}_b$. 

Firstly, $\bm{X}_f\in\mathbb{R}^{H\times W\times C}$ and $\bm{X}_b\in\mathbb{R}^{H\times W\times C}$ are projected to a common space by $f(\cdot)$ and $g(\cdot)$ respectively, where $f(\cdot)$  and $g(\cdot)$ are $1 \times 1$ convolutional layer with spectral normalization~\cite{miyato2018spectral}. For ease of calculation, we reshape $f(\bm{X}_b) \in  \mathbb{R}^{W \times H \times \frac{C}{8}} $ (\emph{resp.}, $g(\bm{X}_f) \in  \mathbb{R}^{W \times H \times \frac{C}{8}} $) into $\bar{f}(\bm{X}_b) \in  \mathbb{R}^{N \times \frac{C}{8}}$ (\emph{resp.}, $\bar{g}(\bm{X}_f) \in  \mathbb{R}^{N \times \frac{C}{8}}$), in which $N=W \times H$. Then, we can calculate the affinity map between $\bm{X}_f$ and $\bm{X}_b$:
\begin{equation}\label{eqn:attention_map}
\begin{aligned}
\bm{A} = softmax\left(\bar{g}(\bm{X}_f)\bar{f}(\bm{X}_b)^{T} \right).
\end{aligned}
\end{equation}
With obtained affinity map $\bm{A}$, we attend information from $\bm{X}_b$ and arrive at the attended feature map $\bm{X}_b'$:
\begin{equation}
\begin{aligned}
\bm{X}_b' =  v\left(\bm{A} \bar{h}(\bm{X}_b)\right),
\end{aligned}
\end{equation}
where $\bar{h}(\cdot)$ means $1 \times 1  $ convolutional layer followed by reshaping to $\mathbb{R}^{N \times \frac{C}{8}}$, similar to $\bar{f}(\cdot)$ and $\bar{g}(\cdot)$ in Eqn.~\ref{eqn:attention_map}. $v(\cdot)$ reshapes the feature map back to $\mathbb{R}^{W \times H \times \frac{C}{8}}$ and then performs $1 \times 1 $ convolution.  Because the attended illumination information should be combined with the foreground information to generate foreground shadow mask, we concatenate $\bm{X}_b'$ and $\bm{X}_{f}$, which is fed into the decoder $D_S$ to produce foreground shadow mask $\tilde{\bm{M}}_{fs}$:
\begin{equation}
\begin{aligned}
\tilde{\bm{M}}_{fs} = D_{S}([\bm{X}_b', \bm{X}_{f}]),
\end{aligned}
\end{equation}
which is enforced to be close to the ground-truth foreground shadow mask $\bm{M}_{fs}$ by
\begin{equation} \label{eqn:loss_reconstruction}
\begin{aligned}
\mathcal{L}_{S} = || \bm{M}_{fs} - \tilde{\bm{M}}_{fs}||_{2}^2.
\end{aligned}
\end{equation}

Although cross attention is not a new idea, this is the first time to achieve foreground-background interaction via cross-attention in shadow generation task.

\begin{figure}
\begin{center}
\includegraphics[scale=0.24]{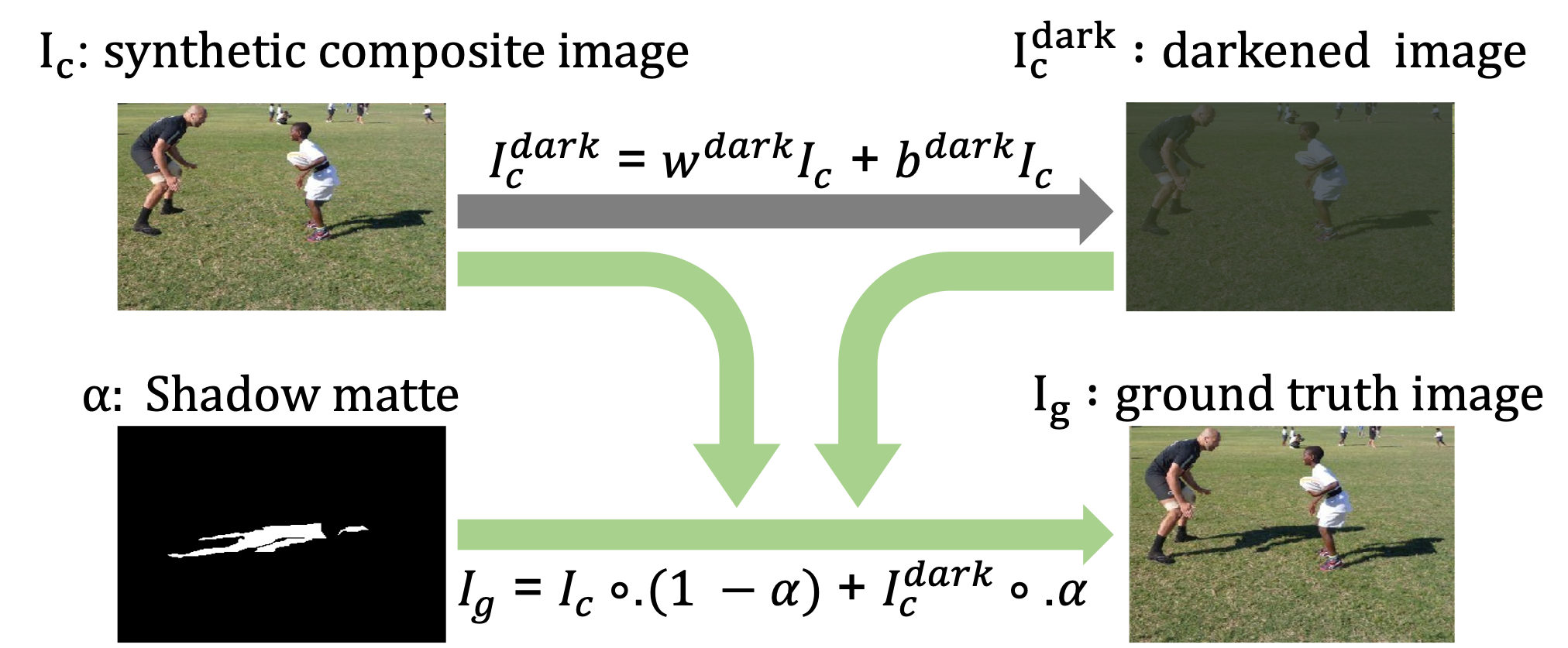}
\end{center}
\caption{Shadow generation via image composition. The ground-truth target image $\mathbf{I}_g$ with foreground shadow can be expressed as the combination of synthetic composite image $\mathbf{I}_c$ and darkened image $\mathbf{I}_c^{dark}$ with shadow matte $\alpha$.}
\label{fig:composition_system} 
\end{figure}

\subsection{Shadow Area Filling}
We design our shadow filling stage based on the illumination model used in ~\cite{shor2008shadow,le2019shadow}. According to \cite{shor2008shadow,le2019shadow}, the value of a shadow-free pixel $I^{lit}(k,i)$ can be linearly transformed from its shadowed value $I^{dark}(k,i)$:
\begin{eqnarray}
I^{lit}(k,i) = w^{lit}(k) I^{dark}(k,i) +b^{lit}(k),
\end{eqnarray}
in which $I(k,i)$ represents the value of the pixel $i$ in color channel $k$ ($k\in R,G,B$). $w^{lit}(k)$ and $b^{lit}(k)$ are constant across all pixels in the umbra area of the shadow. Inversely, the value of a shadowed pixel $I^{dark}(k,i)$ can be linearly transformed from its shadow-free value $I^{lit}(k,i)$:
\begin{eqnarray}
I^{dark}(k,i) = w^{dark}(k) I^{lit}(k,i) +b^{dark}(k).
\end{eqnarray}

To accurately locate the foreground shadow area, we tend to learn a soft shadow matte $\bm{\alpha}$. The value of $\bm{\alpha}$ is 0 in the non-shadow area, 1 in the umbra of shadow area, and varying gradually in the penumbra of shadow area. Then, the target image with foreground shadow can be obtained using the following composition system (see Figure~\ref{fig:composition_system}):
\begin{eqnarray}
\label{eqn:mat_1}
\mathbf{I}_g = \mathbf{I}_c \circ (\mathbf{1} - {\bm{\alpha}}) + \mathbf{I}_c^{dark} \circ  {\bm{\alpha}}, \label{eqn:mat_1_1}\\
\mathbf{I}_c^{dark}(k) = w^{dark}(k) \mathbf{I}_c(k) + b^{dark}(k), \label{eqn:mat_1_2}
\end{eqnarray}
in which $\circ$ means element-wise multiplication, $\mathbf{I}(k)$ represents image $\mathbf{I}$ in color channel $k$,  $\mathbf{I}_c^{dark}(k)$ is the darkened version of $\mathbf{I}_c(k)$ through Eqn.~\ref{eqn:mat_1_2}. $\mathbf{w}^{dark}=[w^{dark}(R), w^{dark}(G), w^{dark}(B)]$ and similarly defined $\mathbf{b}^{dark}$ are called shadow parameters. Given paired images $\{\mathbf{I}_c, \mathbf{I}_g\}$, the ground-truth shadow parameter $\{\mathbf{w}^{dark},\mathbf{b}^{dark}\}$ for the foreground shadow can be easily calculated by using linear regression~\cite{shor2008shadow}. Specifically, we need to calculate the optimal regression coefficients $\{\mathbf{w}^{dark},\mathbf{b}^{dark}\}$ which regress pixel values $\mathbf{I}_c(k,i)$ to $\mathbf{I}_g(k,i)$ in the foreground shadow area. The ground-truth shadow parameters of training images can be precomputed before training, but the ground-truth shadow parameters of test images are unavailable in the testing stage. 
Thus, we learn a shadow parameter predictor $E_{P}$ to estimate $\{\mathbf{w}^{dark},\mathbf{b}^{dark}\}$.

Our $E_{P}$ is implemented as an encoder, which takes the concatenation of composite image ${\mathbf{I}}_c$ and predicted shadow mask $\tilde{\bm{M}}_{fs}$ as input to predict the shadow parameters $\{\tilde{\bm{w}}^{dark}, \tilde{\bm{b}}^{dark}\}$:
\begin{equation}
\begin{aligned}
\{\tilde{\bm{w}}^{dark}, \tilde{\bm{b}}^{dark}\} = E_{P}({\mathbf{I}}_c, \tilde{\bm{M}}_{fs}).
\end{aligned}
\end{equation}
$\{\tilde{\bm{w}}^{dark}, \tilde{\bm{b}}^{dark}\}$ are supervised with ground-truth shadow parameters $\{\bm{w}^{dark}, \bm{b}^{dark}\}$ by regression loss:
\begin{equation} \label{eqn:loss_reconstruction_sp}
\begin{aligned}
\mathcal{L}_{P} = ||\bm{w}^{dark} - \tilde{\bm{w}}^{dark}||_2^2+||\bm{b}^{dark} - \tilde{\bm{b}}^{dark}||_{2}^2.
\end{aligned}
\end{equation}

After estimating $\{\tilde{\bm{w}}^{dark}, \tilde{\bm{b}}^{dark}\}$, we can get the darkened image $\tilde{\mathbf{I}}_c^{dark}(k) = \tilde{w}^{dark}(k) \mathbf{I}_c(k) +\tilde{b}^{dark}(k)$ via Eqn.~\ref{eqn:mat_1_2}. Then, to obtain the final target image, we need to learn a shadow matte $\bm{\alpha}$ for image composition as in Eqn.~\ref{eqn:mat_1}. Our shadow matte generator $G_M$ is based on U-Net~\cite{ronneberger2015u}  with encoder $E_M$ and decoder $D_M$. $G_M$ concatenates composite image $\mathbf{I}_c$, darkened image $\tilde{\mathbf{I}}_c^{dark}$, and predicted shadow mask $\tilde{\bm{M}}_{fs}$ as input, producing the shadow matte $\tilde{\bm{\alpha}}$:
\begin{equation}
\begin{aligned}
\tilde{\bm{\alpha}} = G_M(\mathbf{I}_c, \tilde{\mathbf{I}}_c^{dark}, \tilde{\bm{M}}_{fs}).
\end{aligned}
\end{equation}

Finally, based on $\tilde{\mathbf{I}}_c^{dark}$, $\mathbf{I}_c$, and $\tilde{\bm{\alpha}}$, the target image with foreground shadow can be composed by
\begin{equation} \label{eqn:final_output}
\begin{aligned}
\tilde{\mathbf{I}}_g = \mathbf{I}_c \circ (\mathbf{1} - \tilde{\bm{\alpha}}) + \tilde{\mathbf{I}}_c^{dark} \circ \tilde{\bm{\alpha}}.
\end{aligned}
\end{equation}
The generated target image is supervised by the ground-truth target image with a reconstruction loss:
\begin{equation} \label{eqn:loss_reconstruction_image}
\begin{aligned}
\mathcal{L}_{I} = || \mathbf{I}_g - \tilde{\mathbf{I}}_g||_{2}^2.
\end{aligned}
\end{equation}

To the best of our knowledge, we are the first to generate shadow by blending original image and darkened image. 

\subsection{Conditional Discriminator}
To ensure that the generated shadow mask $\tilde{\bm{M}}_{fs}$ and the generated target image $\tilde{\mathbf{I}}_g$ are close to real shadow mask $\bm{M}_{fs}$ and real target image $\mathbf{I}_g$ respectively, we design a conditional discriminator $D$ to bridge the gap between the generated triplet $\{\tilde{\bm{M}}_{fs}, \tilde{\mathbf{I}}_g, \bm{M}_{fo}\}$ and the real triplet $\{\bm{{M}}_{fs},\mathbf{I}_g, \bm{M}_{fo}\}$. The architecture of our conditional discriminator is similar to Patch-GAN~\cite{pix2pix2017}, which takes the concatenation of triplet as input. We adopt the hinge adversarial loss~\cite{miyato2018cgans} as follows,
\begin{eqnarray} \label{eqn:adversarial_delta_loss}
\!\!\!\!\!\!\!\!&&\mathcal{L}_D = \mathbb{E}_{ \tilde{\bm{M}}_{fs}, \tilde{\mathbf{I}}_g, \bm{M}_{fo} } [\max (0,1 + \mathrm{D}(\tilde{\bm{M}}_{fs}, \tilde{\mathbf{I}}_g, \bm{M}_{fo}))]  +  \nonumber\\
\!\!\!\!\!\!\!\!&&\quad\quad\mathbb{E}_{\bm{{M}}_{fs}, \mathbf{I}_g, \bm{M}_{fo} }  [\max(0, 1 - \mathrm{D}(\bm{{M}}_{fs}, \mathbf{I}_g, \bm{M}_{fo})], \nonumber\\
\!\!\!\!\!\!\!\!&&\mathcal{L}_{GD} = - \mathbb{E}_{\tilde{\bm{M}}_{fs}, \tilde{\mathbf{I}}_g, \bm{M}_{fo}} [\mathrm{D}(\tilde{\bm{M}}_{fs}, \tilde{\mathbf{I}}_g, \bm{M}_{fo})].
\end{eqnarray}

\setlength{\tabcolsep}{4.5pt}
\begin{table*}[t]
  \begin{center}
  \begin{tabular}{l|cccc|cccc}
      \toprule[0.8pt]
      \multirow{2}{*}{Method}
      &\multicolumn{4}{c|}{BOS Test Images}
      &\multicolumn{4}{c}{BOS-free Test Images}\cr 
      &GRMSE $\downarrow$ &LRMSE $\downarrow$  &GSSIM $\uparrow$  &LSSIM $\uparrow$ &GRMSE $\downarrow$ &LRMSE $\downarrow$  &GSSIM $\uparrow$  &LSSIM $\uparrow$ \cr
      \cmidrule(r){1-1} 
      \cmidrule(r){2-5}  
      \cmidrule(r){6-9} 
        Pix2Pix &  7.659&	75.346& 0.926&0.249	&18.875	&81.444	&0.858 &0.110	 \cr
    Pix2Pix-Res   & 5.961&	76.046&	0.971&0.253 	&18.365	&81.966	&0.901 &0.107
    \cr
    ShadowGAN  & 5.985&	78.412&	0.984& 0.240 & 19.306&	87.017&	0.918& 0.078	\cr
    Mask-ShadowGAN  &8.287	&79.212& 0.952&0.245 & 19.475	&83.457  &0.891&0.109	  \cr
    ARShadowGAN   &6.481&	75.099&	0.983 & 0.251	&18.723	&81.272 &0.917	& 0.109 \cr
    Ours    &\textbf{4.754}&	\textbf{61.763}&	\textbf{0.988}&  \textbf{0.380}	  &\textbf{15.128}	&\textbf{61.439}	&\textbf{0.928} &\textbf{0.183} \cr
    \bottomrule[0.8pt]
    \end{tabular}
    \end{center}
   \caption{Results of quantitative comparison on our DESOBA dataset. }
     \label{tab:metric_compare}
\end{table*}

\subsection{Optimization}
The overall optimization function can be written as
\begin{equation}
\label{eqn:optimization_total}
\begin{aligned}
\mathcal{L} =  \lambda_S \mathcal{L}_{S} +  \lambda_I \mathcal{L}_{I} +  \lambda_P \mathcal{L}_{P} + \lambda_{GD} \mathcal{L}_{GD} +  \mathcal{L}_{D},
\end{aligned}
\end{equation}
where $\lambda_S$, $\lambda_I$, $\lambda_P$, and $\lambda_{GD}$ are trade-off parameters. 

The parameters of $\{E_S, CAI, D_S, E_P, E_M, D_M\}$ are denoted as $\theta_G$, while the parameters of $D$ are denoted as $\theta_D$. Following adversarial learning framework~\cite{gulrajani2017improved}, we use related loss terms to optimize  $\theta_G$ and  $\theta_D$ alternatingly. In detail, $\theta_D$ is optimized by minimizing $\mathcal{L}_{D}$. Then, $\theta_G$ is optimized by minimizing $ \lambda_S \mathcal{L}_{S} +  \lambda_I \mathcal{L}_{I} +  \lambda_P \mathcal{L}_{P} + \lambda_{GD} \mathcal{L}_{GD}$.

\section{Experiments}\label{sec:experiments}
\subsection{Experimental Setup}  \label{sec:implementation}
\noindent\textbf{Datasets}
We conduct experiments on our constructed DESOBA dataset and real composite images. 
On DESOBA dataset, we perform both quantitative and qualitative evaluation based on $615$ test image pairs with one foreground, which are divided into $581$ BOS test image pairs and $34$ BOS-free test image pairs. We also show the qualitative results of test images with two foregrounds in Supplementary.
\textit{The experiments on real composite images will be described in Supplementary due to space limitation.} 

\noindent\textbf{Implementation}
After a few trials, we set $\lambda_S =\lambda_I = 10$, $\lambda_P = 1$, and $\lambda_{GD} = 0.1$ by observing the generated images during training.  We use Pytorch $1.3.0$ to implement our model, which is distributed on RTX 2080 Ti GPU.
All images in our used datasets are resized to $256 \times 256$ for training and testing. We use adam optimizer with the learning rate initialized as $0.0002$ and $\beta$ set to $(0.5, 0.99)$. The batch size is  $1$ and our model is trained for $50$ epochs.

\noindent\textbf{Baselines}
Following~\cite{liu2020arshadowgan}, we select Pix2Pix~\cite{pix2pix2017}, Pix2Pix-Res, ShadowGAN~\cite{zhang2019shadowgan}, ARShadowGAN~\cite{liu2020arshadowgan}, and Mask-ShadowGAN~\cite{hu2019mask} as baselines. Pix2Pix~\cite{pix2pix2017} is a popular image-to-image translation method, which takes composite image as input and outputs target image. Pix2Pix-Res has the same architecture as Pix2Pix except producing a residual image, which is added to the input image to generate the target image. ShadowGAN~\cite{zhang2019shadowgan} and ARShadowGAN~\cite{liu2020arshadowgan} are two closely related methods, which can be directly applied to our task. Mask-ShadowGAN~\cite{hu2019mask} originally performs both mask-free shadow removal and mask-guided shadow generation. We adapt it to our task by exchanging two generators to perform mask-guided shadow removal and mask-free shadow generation, in which the mask-free shadow generator can be used in our task.

\noindent\textbf{Evaluation Metrics}
\label{sec:metric}
Following~\cite{liu2020arshadowgan}, we adopt Root Mean Square Error (RMSE) and Structural SIMilarity index (SSIM).
RMSE and SSIM are calculated based on the ground-truth target image and the generated target image.
Global RMSE (GRMSE) and Global SSIM (GSSIM) are calculated over the whole image, while Local RMSE (LRMSE) and Local SSIM (LSSIM) are calculated over the ground-truth foreground shadow area. 

\subsection{Evaluation on Our DESOBA Dataset} \label{sec:visualization}
On DESOBA dataset, BOS test set and BOS-free test set are evaluated separately and the comparison results are summarized in Table~\ref{tab:metric_compare}. We can observe that our SGRNet achieves the lowest GRMSE, LRMSE and the highest GSSIM, LSSIM, which demonstrates that our method could generate more realistic and compatible shadows for foreground objects compared with baselines. 
The difference between the results on BOS test set and  BOS-free test set is partially caused by the size of foreground shadow, because BOS-free test images usually have larger foreground shadows than BOS test images as shown in Figure~\ref{fig:dataset_samples}. \emph{We will provide more in-depth comparison by controlling the foreground shadow size in Supplementary.}



\begin{figure*}
\begin{center}
\includegraphics[scale=0.59]{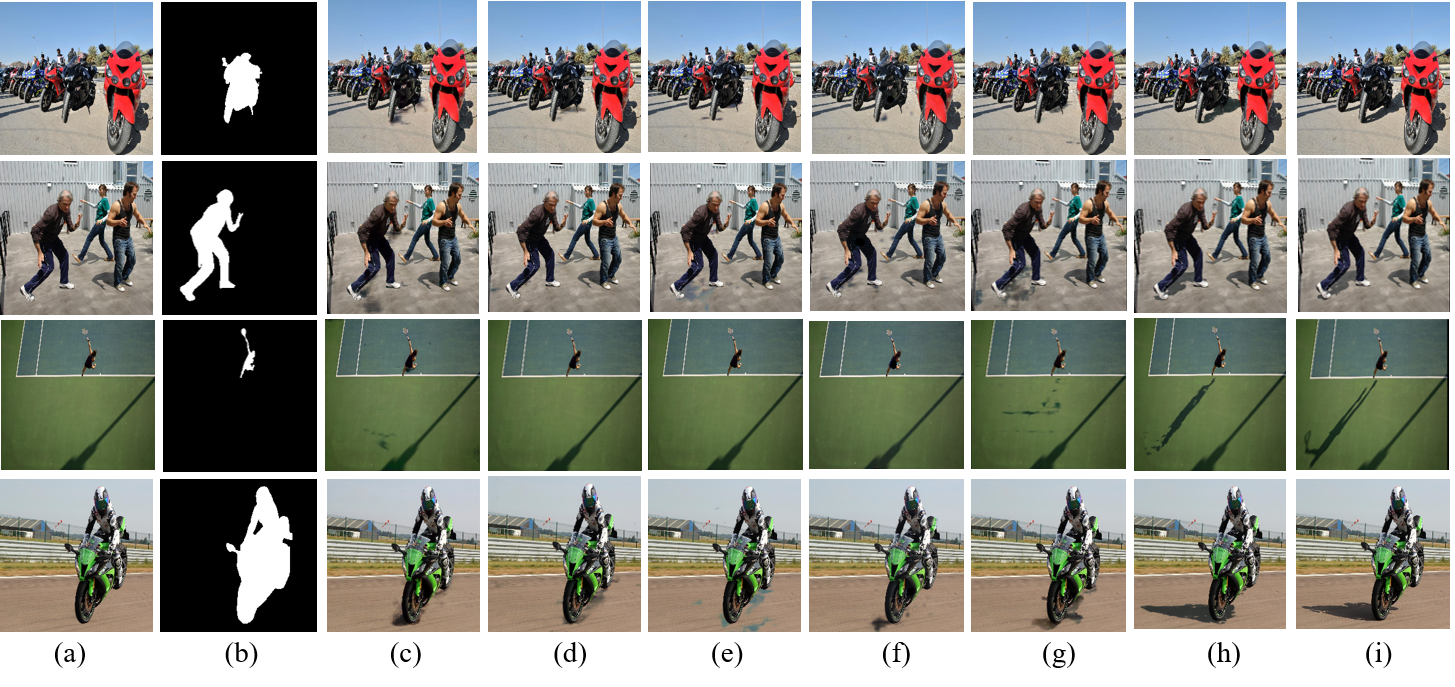}
\end{center}
\caption{Visualization comparison on our DESOBA dataset. From left to right are input composite image (a), foreground object mask (b), results of Pix2Pix
(c), Pix2Pix-Res (d), ShadowGAN (e), Mask-ShadowGAN (f), ARShadowGAN (g), our SGRNet (h), ground-truth (i). The results on BOS (\emph{resp.,}  BOS-free) test images are shown in row 1-2 (\emph{resp.,} 3-4).}
\label{fig:compare_images} 
\end{figure*}

For qualitative comparison, we show some example images generated by our SGRNet and other baselines on BOS and BOS-free test images in Figure~\ref{fig:compare_images}. We can see that our SGRNet can generally generate foreground shadows with reasonable shapes and shadow directions compatible with the object-shadow pairs in background. In contrast, other baselines produce foreground shadows with implausible shapes, or even fail to produce any shadow. Our method can also generate reasonable shadows for BOS-free test images, because the background in BOS-free images could also provide a set of illumination cues (\emph{e.g.}, shading, sky appearance variation)~\cite{lalonde2012estimating, zhang2019all} as discussed in Section~\ref{sec:dataset_cons}. \emph{More visualization results including the intermediate results (e.g., generated foreground shadow mask, generated darkened image) can be found in Supplementary.} 

\setlength{\tabcolsep}{4pt}
\begin{table}[t]
  \begin{center}
  \begin{tabular}{l|rrrr}
      \toprule[0.8pt]
      Method &GRMSE $\downarrow$ &LRMSE $\downarrow$  &GSSIM $\uparrow$  &LSSIM$\uparrow$ \cr
    \hline
    
        w/o $E_{BS}$  & 5.549	&68.876& 	0.985&0.317 \cr
    w/o CAI &5.106	&68.031	&0.986 &0.320	 \cr
    w/o $\mathbf{M}_{bos}$ &4.931	&63.141	&0.986 &0.358	 \cr	
     \hline
    w/o Fill & 5.328	&67.789 & 0.941 &0.255		 \cr
    w/o $\mathcal{L}_{P}$ &4.929  &65.054 &0.986 &0.352 \cr
     \hline
    Naive D & 5.059& 65.238& 0.987& 0.355 \cr
    w/o $\mathcal{L}_{GD}$ & 5.453& 67.056 & 0.986& 0.348 \cr
     Ours  &\textbf{4.754}&	\textbf{61.763}&	\textbf{0.988}&  \textbf{0.380}  \cr
    \bottomrule[0.8pt]
    \end{tabular}
    \end{center}
  \caption{Ablation studies of loss terms and alternative network designs on BOS test images from DESOBA dataset. }
    \label{tab:ablation}
\end{table}

\subsection{Ablation Studies}
We analyze the impact of loss terms and alternative network designs of our SGRNet on BOS test images from DESOBA dataset. Quantitative results are reported in Table~\ref{tab:ablation}. 

\noindent\textbf{Shadow mask prediction stage}: To investigate the necessity of background encoder, we remove the background encoder $E_{BS}$, which is referred to as ``w/o $E_{BS}$" in Table~\ref{tab:ablation}. To verify the effectiveness of Cross-Attention Integration (CAI) layer, we remove CAI layer and directly concatenate $[\bm{X}_f,\bm{X}_b]$, which is referred to as ``w/o CAI". The performance of ``w/o CAI" is better than ``w/o $E_{BS}$", which shows the advantage of extracting foreground and background information separately.
The performance of ``w/o CAI" is worse than our full method, which shows the benefit of encouraging thorough information interaction between foreground and background.
To study the importance of background object-shadow mask, we set the value of $\mathbf{M}_{bos}$ as zero, which is referred to as  ``w/o $\mathbf{M}_{bos}$". The performance is better than ``w/o $E_{BS}$" and ``w/o CAI", which can be explained as follows. CAI layer can help foreground encoder exploit illumination information from background, even without explicit background object-shadow mask. The comparison between ``w/o $\mathbf{M}_{bos}$" and full method proves that background object-shadow mask can indeed provide useful shadow cues as guidance.

\noindent\textbf{Shadow filling stage}: To corroborate the superiority of image composition system in Section~\ref{sec:method}, we replace our $E_{P}$ and $\{E_M, D_M\}$ with a U-Net~\cite{ronneberger2015u} model which takes  $\tilde{\mathbf{M}}_{fs}$ and $\mathbf{I}_c$ as input to generate the final target image directly, which is referred to as ``w/o Fill" in Table~\ref{tab:ablation}. The result is worse than full method, which demonstrates the advantage of composition system. We also remove the supervision for shadow parameters by setting $\mathcal{L}_{P} = 0$, which is referred to as ``w/o $\mathcal{L}_{P}$". We find that the performance is better than ``w/o Fill" but worse than full method, which demonstrates the necessity of supervision from ground-truth shadow parameters.

\noindent\textbf{Adversarial learning}: We remove conditional information $\{\bm{M}_{fo}, \bm{M}_{fs}\}$ (\emph{resp.,}  $\{\bm{M}_{fo}, \tilde{\bm{M}}_{fs}\}$), and only feed $\mathbf{I}_g$ (\emph{resp.,} $\tilde{\mathbf{I}}_g$) into the discriminator $D$, which is named as ``Naive D" in Table~\ref{tab:ablation}. It can be seen that conditional discriminator can enhance the quality of generated images. To further investigate the effect of adversarial learning, we remove the adversarial loss  $\mathcal{L}_{GD}$ from Eqn.~\ref{eqn:optimization_total} and report the result as ``w/o $\mathcal{L}_{GD}$". The result is worse than ``Naive D", which indicates that adversarial learning can help generate more realistic foreground shadows.

We visualize some examples produced by different ablated methods and conduct ablation studies on BOS-free test images in Supplementary.

\subsection{Evaluation on Real Composite Images}

To obtain real composite images, we select test images from DESOBA as background images, and paste foreground objects also from test images at reasonable locations on the background images. In this way, we create $100$ real composite images without foreground shadows for evaluation.
Because real composite images do not have ground-truth target images, it is impossible to perform quantitative evaluation. Therefore, we conduct user study on the $100$ composite images for subjective evaluation. The visualization results and user study are left to Supplementary.

\section{Conclusion}
In this work, we have contributed a real-world shadow generation dataset DESOBA. We have also proposed SGRNet, a novel shadow generation method, which can predict shadow mask by inferring illumination information from background and estimate shadow parameters based on illumination model. The promising results on our constructed dataset and real composite images have demonstrated the effectiveness of our method.


\section*{Acknowledgements}
 This work is partially sponsored by National Natural Science Foundation of China (Grant No. 61902247),  Shanghai Municipal Science and Technology Major Project (2021SHZDZX0102), Shanghai Municipal Science and Technology Key Project (Grant No. 20511100300).

\nolinenumbers
{
\bibliography{egbib.bib}
} 

\end{document}


\nolinenumbers
\maketitle

In this document, we provide additional materials to support our main submission. In Section~\ref{sec:architecture}, we describe the structure of our shadow mask generator $G_S$, shadow parameter predictor $E_P$, shadow matte predictor $G_M$, and discriminator $D$. We also compare the model size between our SGRNet and baseline methods. 
In Section~\ref{sec:evaluation_real_composite}, we evaluate on the real composite images. 
In Section~\ref{sec:immpact_shadow_size}, we report the comparison results according to the foreground shadow ratio (\emph{i.e.}, the area of ground-truth foreground shadow over the area of the whole image). 
In Section \ref{sec:vis_ablated_methods}, we show some example images produced by the ablated versions of our method on BOS test images from DESOBA dataset.
In Section~\ref{sec:ablated_methods_bosfree}, we report the results of ablated methods on BOS-free images from DESOBA dataset.
In Section~\ref{sec:vis_intermediate_results}, we show some intermediate results produced by our SGRNet in each step. 
In Section~\ref{sec:more_results_on_desoba}, we visualize more images generated from our SGRNet on DESOBA dataset.
In Section~\ref{sec:two_foreground_object}, we visualize some example images with two foreground objects on DESOBA dataset.
In Section~\ref{sec:failure_case}, we show some failure cases generated by our SGRNet. 


\setlength{\tabcolsep}{6pt}
\begin{table}[t]
\begin{center}
  \begin{tabular}{ccccc}
  \toprule[0.8pt]
    Module & Layer & Resample  & Output \cr
      
    \hline
     & $\{\mathbf{I}_c,\bm{M}\}$ & -    & 256$\times$256$\times$5   \cr
     \hline
     &Conv  &  -  &   256$\times$256$\times$32  \cr
     \hline
     $E_{FS}$ / $E_{BS}$ &DBlk  &  AvgPool    &  128$\times$128$\times$64  \cr
     &DBlk  &   AvgPool   & 64$\times$64$\times$128  \cr
     &DBlk  &   AvgPool  &  32$\times$32$\times$256  \cr
     &DBlk  &   AvgPool   &  16$\times$16$\times$512  \cr
     
     \hline
     $CAI$   &Conv & - & 16$\times$16$\times$512  \cr
     &Conv & - & 16$\times$16$\times$512  \cr
     &Conv & - & 16$\times$16$\times$512  \cr
     &Conv & - & 16$\times$16$\times$512  \cr
     \hline
     $D_S$ &UBlk  &   Upsample   &   32$\times$32$\times$512 \cr
     &UBlk  &  Upsample   &  64$\times$64$\times$256  \cr
     &UBlk  & Upsample   &   128$\times$128$\times$128 \cr
     &UBlk  &  Upsample   & 256$\times$256$\times$64   \cr
     \hline
     &Conv  & - &   256$\times$256$\times$1 \cr
    \bottomrule[0.8pt]   
  \end{tabular}
  \end{center} 
  \caption{The network architecture of our shadow mask generator $G_S$ consisting of a foreground encoder $E_{FS}$, a background encoder $E_{BS}$, a Cross-Attention Integration (CAI) layer, and a decoder $D_S$. $\bm{I}_c$ is the input composite image. $\bm{M} = \{\bm{M}_{fo}, \bm{M}_{bos} \}$ consists of a foreground shadow mask $\bm{M}_{fo}$ and the mask of object-shadow pairs $\bm{M}_{bos}$ in background.} 
  \label{tab:msg}
\end{table}

\setlength{\tabcolsep}{6pt}
\begin{table}[t]
\begin{center}
  \begin{tabular}{ccccc}
  \toprule[0.8pt]
    Module&Layer & Resample & Output \cr
      
    \hline
     & $\{\mathbf{I}_c, \tilde{\bm{M}}_{fs}\}$  & -   &   256$\times$256$\times$4   \cr
     \hline
    $E_P$ &DBlk  &  MaxPool  &  128$\times$128$\times$32  \cr
      &DBlk  &   MaxPool  &   64$\times$64$\times$64  \cr
      &DBlk  &   MaxPool  &   32$\times$32$\times$128  \cr
     &DBlk  &   MaxPool  &   16$\times$16$\times$256  \cr
     & GAP & - & 1$\times$256 \cr
     &FC  &  - & 1$\times$6  \cr
      \bottomrule[0.8pt]
  \end{tabular}
\end{center}
  \caption{The network architecture of shadow parameter predictor $E_{P}$.} 
    \label{tab:SP}
\end{table}

\setlength{\tabcolsep}{6pt}
\begin{table}[t]
\begin{center}
  \begin{tabular}{ccccc}
  \toprule[0.8pt]
    Module&Layer & Resample & Output \cr
      
    \hline
     & $(\tilde{\mathbf{I}}$ / $\mathbf{I})$  & -   &   256$\times$256$\times$5   \cr
     \hline
    $D$ &DBlk  &  AvgPool  &  128$\times$128$\times$64  \cr
      &DBlk  &   AvgPool  &   64$\times$64$\times$128  \cr
      &DBlk  &   AvgPool  &   32$\times$32$\times$256  \cr
     &DBlk  &   AvgPool  &   16$\times$16$\times$512  \cr
     & Conv & - & 16$\times$16$\times$1 \cr
      \bottomrule[0.8pt]
  \end{tabular}
      \end{center}
    \caption{The network architecture of discriminator $D$. $\tilde{\mathbf{I}} = \{ \tilde{\bm{M}}_{fs}, \tilde{\mathbf{I}}_g, \mathbf{M}_{fo}\}$ (\emph{resp.,} $\mathbf{I} = \{\bm{M}_{fs}, \mathbf{I}_g, \mathbf{M}_{fo}\}$) denotes the fake (\emph{resp.,} real) triplet of discriminator.} 
      \label{tab:d}
\end{table}

\setlength{\tabcolsep}{4pt}
\begin{table}[t]
  
        \begin{center}

  \begin{tabular}{lrr}  
    \toprule[0.8pt]
     Setting& Training phase   & Testing phase \cr
    \hline
     Pix2Pix & 11.9M & 5.9M  \cr
    \hline
    Pix2Pix-Res & 11.9M & 5.9M \cr
    \hline
    ShadowGAN &21.8M & 9.6M  \cr
    \hline
    Mask-ShadowGAN &32.5M &  27.1M \cr
    \hline
    ARShadowGAN &21.8M & 14.6M  \cr
    \hline
    SGRNet & 18.1M & 11.2M   \cr
    \bottomrule[0.8pt]
  \end{tabular}
        \end{center}

  \caption{Model size comparison among  Pix2Pix~\cite{pix2pix2017}, Pix2Pix-Res, ShadowGAN~\cite{zhang2019shadowgan}, Mask-ShadowGAN~\cite{hu2019mask}, ARShadowGAN~\cite{liu2020arshadowgan}, and our SGRNet.} 
    \label{tab:model_parameters}
\end{table}

\section{Details of Network Architecture}\label{sec:architecture}
\noindent\textbf{Shadow Mask Generator} Our shadow mask generator $G_S$ based on U-Net structure~\cite{ronneberger2015u} consists of a foreground encoder $E_{FS}$, a background encoder $E_{BS}$, a Cross-Attention Integration (CAI) layer, and a decoder $D_S$. Our foreground encoder $E_{FS}$ has $4$ downsampling blocks (DBlk), in which each downsampling block consists of $1$ convolutional layer with ReLU and batch normalization followed by one downsampling layer. The structure of background encoder $E_{BS}$ is the same as foreground encoder $E_{FS}$ without parameter sharing. Our CAI layer is constructed by four $1 \times 1$ convolutional layers. 
Our decoder $D_S$ has $4$ upsampling blocks (UBlk), in which each upsampling block is constructed by  $1$ convolutional layer with ReLU and batch normalization followed by one upsampling layer. The architecture of our shadow mask generator $G_S$ is summarized in Table~\ref{tab:msg}.

\noindent\textbf{Shadow Parameter Predictor} Our shadow parameter predictor $E_{P}$ is constructed by $4$ downsampling blocks (DBlk), $1$ Global Average Pooling layer (GAP), and $1$ fully-connected (FC) layer, in which each DBlk consists of  $1$ convolutional layer with ReLU followed by one downsampling layer. The architecture of our $E_{P}$ is summarized in Table~\ref{tab:SP}.

\noindent\textbf{Shadow Matte Generator}
Our shadow matte generator $G_{M}$ is also constructed by U-Net ~\cite{ronneberger2015u} with an encoder $E_M$ and a decoder $D_M$. The structure of $E_M$ (\emph{resp.,} $D_M$) is the same as $E_{FS}$ (\emph{resp.,} $D_S$) in Table~\ref{tab:msg}.

\noindent\textbf{Conditional Discriminator} To push the generated shadow mask $\tilde{\bm{M}}_{fs}$ and the generated target image $\tilde{\bm{I}}_g$ close to real shadow mask $\bm{M}_{fs}$ and real target image $\mathbf{I}_g$, we use conditional discriminator $D$ to distinguish the generated triplet $\{\tilde{\bm{M}}_{fs}, \tilde{\mathbf{I}}_g, \bm{M}_{fo}\}$ from the real triplet $\{\bm{{M}}_{fs},\mathbf{I}_g, \bm{M}_{fo}\}$. The structure of our $D$ is based on patch-GAN~\cite{pix2pix2017}, which consists of $4$ downsampling blocks (DBlk). Each DBlk has one convolution with valid padding, instance normalization, and LeakyReLU. Then, a convolution produces the last feature map which is activated by sigmoid function. The architecture of our discriminator $D$ is summarized in Table~\ref{tab:d}.

\noindent\textbf{The Number of Model Parameters}
We compare the number of model parameters of our SGRNet with Pix2Pix~\cite{pix2pix2017}, Pix2Pix-Res, ShadowGAN~\cite{zhang2019shadowgan}, Mask-ShadowGAN~\cite{hu2019mask}, and ARShadowGAN~\cite{liu2020arshadowgan} in the training stage and testing stage, respectively. In the training stage, The model parameters of generator and discriminator are trainable to complete two-player adversarial learning, while only generator is used to generate target images in the testing stage.  In Table~\ref{tab:model_parameters}, we can see that our SGRNet uses fewer model parameters to generate images of better quality compared with Mask-ShadowGAN~\cite{hu2019mask} and ARShadowGAN~\cite{liu2020arshadowgan}. 


\begin{figure*}
\begin{center}
\includegraphics[scale=0.8]{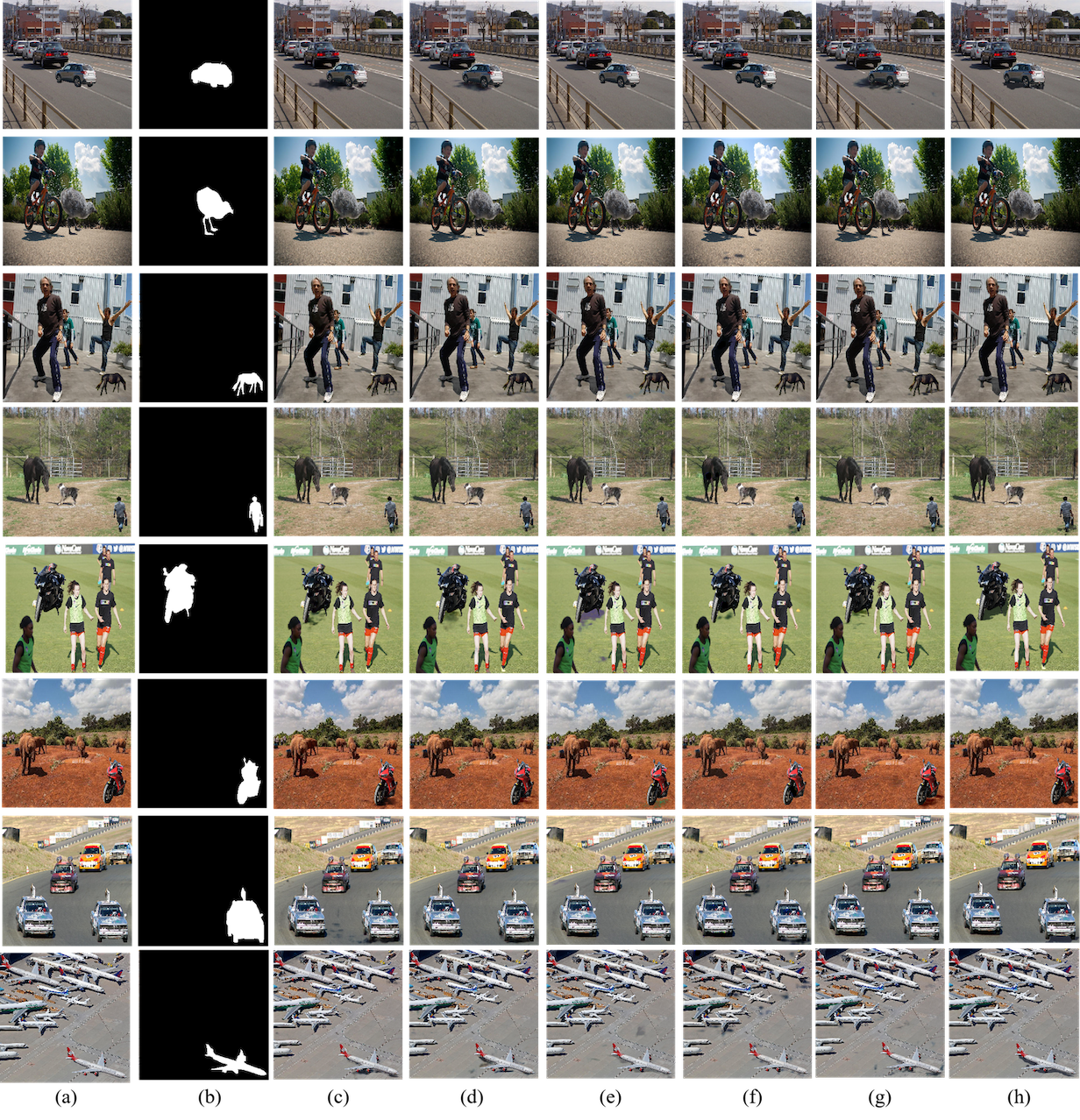}
\end{center}
\caption{Visualization comparison with different methods on real composite images.  From left to right are  input composite image (a), foreground object mask (b), results of Pix2Pix (c), Pix2Pix-Res (d), ShadowGAN (e), Mask-ShadowGAN (f), ARShadowGAN (g), SGRNet (h).}
\label{fig:composite} 
\end{figure*}

\section{Evaluation on Real Composite Images}  \label{sec:evaluation_real_composite}
Since training/test images in DESOBA are synthetic composite images, we further evaluate the effectiveness of our model on real composite images in this section. To obtain real composite images, we select test images from DESOBA as background images, and paste foreground objects also from test images at reasonable locations on the background images. In this way, we create $100$ real composite images without foreground shadows for evaluation.
The six methods in  Table~\ref{tab:model_parameters} 
are used to generate foreground shadows for composite images. 

We visualize the comparison results in Figure~\ref{fig:composite}.
Note that the inserted object (\emph{e.g.}, car, airplane) in row 7-8 is from the original background image, which is a special case of image composition.
We can see that our SGRNet can generate reasonable foreground shadows for foreground objects, which are compatible with the background object-shadow pairs. 

Because real composite images do not have ground-truth target images, it is impossible to perform quantitative evaluation. Therefore, we conduct user study on the $100$ composite images for subjective evaluation. In detail, for each composite image, we can obtain a group of  $7$ images including the input composite images and the generated images from $6$ methods. We construct pairs within each group of $7$ images. Then, we invite $20$ human raters to observe a pair of images at a time and ask him/her to choose the one with more realistic foreground shadow. Finally, we use the Bradley-Terry (B-T) model \cite{bradley1952rank} to calculate the global ranking score for each method, which is reported in Table \ref{tab:bt_score}. 
One observation is that the B-T scores of Pix2Pix~\cite{pix2pix2017}, ShadowGAN~\cite{zhang2019shadowgan}, and Mask-ShadowGAN~\cite{hu2019mask} are even worse than that of input composite images, which
can be explained as follows. Shadows generated by Pix2Pix~\cite{pix2pix2017},  ShadowGAN~\cite{zhang2019shadowgan}, and Mask-ShadowGAN~\cite{hu2019mask} often have incomplete and unreasonable shapes incompatible with background information, making the generated target image even more unrealistic than the input composite image. In contrast, our SGRNet achieves the highest B-T score among all the methods.

\setlength{\tabcolsep}{10pt}
\begin{table}[t]
      \begin{center}

  \begin{tabular}{l r}
      \toprule[0.8pt]
      Method & B-T score $\uparrow$\cr
    \hline
    Input Composite& 0.612 \cr
    Pix2Pix&0.587 \cr
    Pix2Pix-Res& 0.708  \cr
    ShadowGAN& -0.539 \cr
    Mask-ShadowGAN& -0.232  \cr
    ARShadowGAN&0.701   \cr
    SGRNet& 2.509 \cr
    \bottomrule[0.8pt]
    \end{tabular}
      \end{center}

    \caption{B-T scores of different methods on $100$ real composite images.} 
  \label{tab:bt_score}
\end{table}

\setlength{\tabcolsep}{3pt}
\begin{table*}[t]
      \begin{center}

 \begin{tabular}{l|ccccc|ccccc}
      \toprule[0.8pt]
      \multirow{2}{*}{Subsets}
    &\multicolumn{5}{c|}{BOS Test Images}
      &\multicolumn{5}{c}{BOS-free Test Images}\cr 
      & NUM &GRMSE $\downarrow$ &LRMSE $\downarrow$  &GSSIM $\uparrow$   &LSSIM $\uparrow$  & NUM & GRMSE $\downarrow$ &LRMSE $\downarrow$  &GSSIM $\uparrow$  &LSSIM$\uparrow$ \cr
      \cmidrule(r){1-1} 
      \cmidrule(r){2-6}  
      \cmidrule(r){7-11} 
      
    $r_{fs} \in (0,0.02]$ &526 &  3.879	&62.723	 &0.993  &0.395 &	9  & 8.172&	64.468&	0.982& 0.282	 	 \cr
    $r_{fs} \in (0.02,0.04]$  &32 & 10.080	&46.204	 &0.964  &0.216&   8 &14.723&	68.969& 0.942&	0.178		 \cr
    $r_{fs} \in (0.04,0.08]$ & 16 &14.649 &	57.148 & 0.939 &0.153		&  12  &	14.422&	52.084&	0.928 & 0.134		 \cr
    $r_{fs} \in (0.08,1]$ &7 & 23.552	&71.246	 &0.893  &0.145&	5 & 29.990	&66.390& 0.806&0.126	 \cr
    \hline
    $r_{fs} \in (0,1]$  & 581& 4.754	&61.763	 &0.988 & 0.380 & 34& 15.128 &61.439 & 0.928 &0.183	 	 \cr
    \bottomrule[0.8pt]
    \end{tabular}
      \end{center}
    \caption{The impact of shadow size on BOS (\emph{resp.,} BOS-free) test images from DESOBA dataset. NUM denotes the number of BOS (\emph{resp.,} BOS-free) test images in each subset.}
      \label{tab:shadow_size}
\end{table*}

\begin{figure*}[t]
\begin{center}
\includegraphics[scale=0.80]{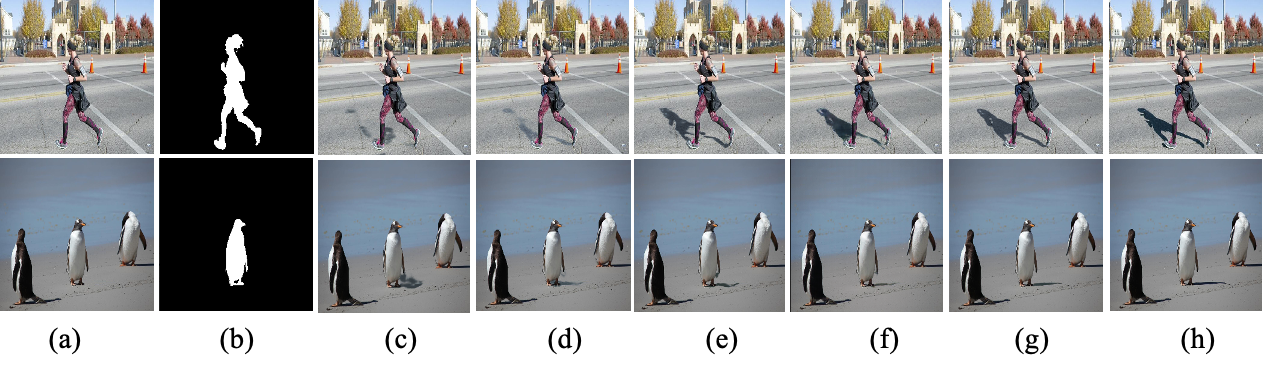}
\end{center}
\caption{Visualization results of ablated versions of our SGRNet on DESOBA dataset. From left to right are input composite image (a), foreground object mask (b), results of w/o $E_{BS}$ (c),
w/o CAI (d), w/o $\mathbf{M}_{bos}$ (e) , w/o Fill (f), SGRNet (g), ground-truth (h). }
\label{fig:abalation} 
\end{figure*}

\setlength{\tabcolsep}{4pt}
\begin{table}[t]
  \begin{center}
  \begin{tabular}{l|rrrrr}
      \toprule[0.8pt]
      Method & GRMSE $\downarrow$ &LRMSE $\downarrow$  &GSSIM $\uparrow$  &LSSIM $\uparrow$ \cr
    \hline
    
    w/o $E_{BS}$  &18.519 	&69.317& 0.914	&0.130 \cr
    w/o CAI& 18.011	&71.029& 0.916	&0.120	 \cr
    w/o $\mathbf{M}_{bos}$& 16.612	&68.487&0.916 	&0.130\cr	
     \hline
    w/o Fill & 16.453	&69.513& 0.897	&0.131	 \cr
    w/o $\mathcal{L}_{P}$& 16.523	&70.248&0.908 	&0.124	 \cr
     \hline
    Naive D &16.213 	&68.864& 0.917	&0.139	 \cr
    w/o $\mathcal{L}_{GD}$ & 17.512	&67.546&0.916 	&0.132	 \cr
     Ours  &\textbf{15.128}&	\textbf{61.439}&	\textbf{0.928}&\textbf{0.183}  \cr
    \bottomrule[0.8pt]
    \end{tabular}
    \end{center}
  \caption{Ablation studies of loss terms and alternative network designs on BOS-free test images from DESOBA dataset. }
    \label{tab:ablation_bosfree}
\end{table}

\begin{figure*} [t]
\begin{center}
\includegraphics[scale=0.44]{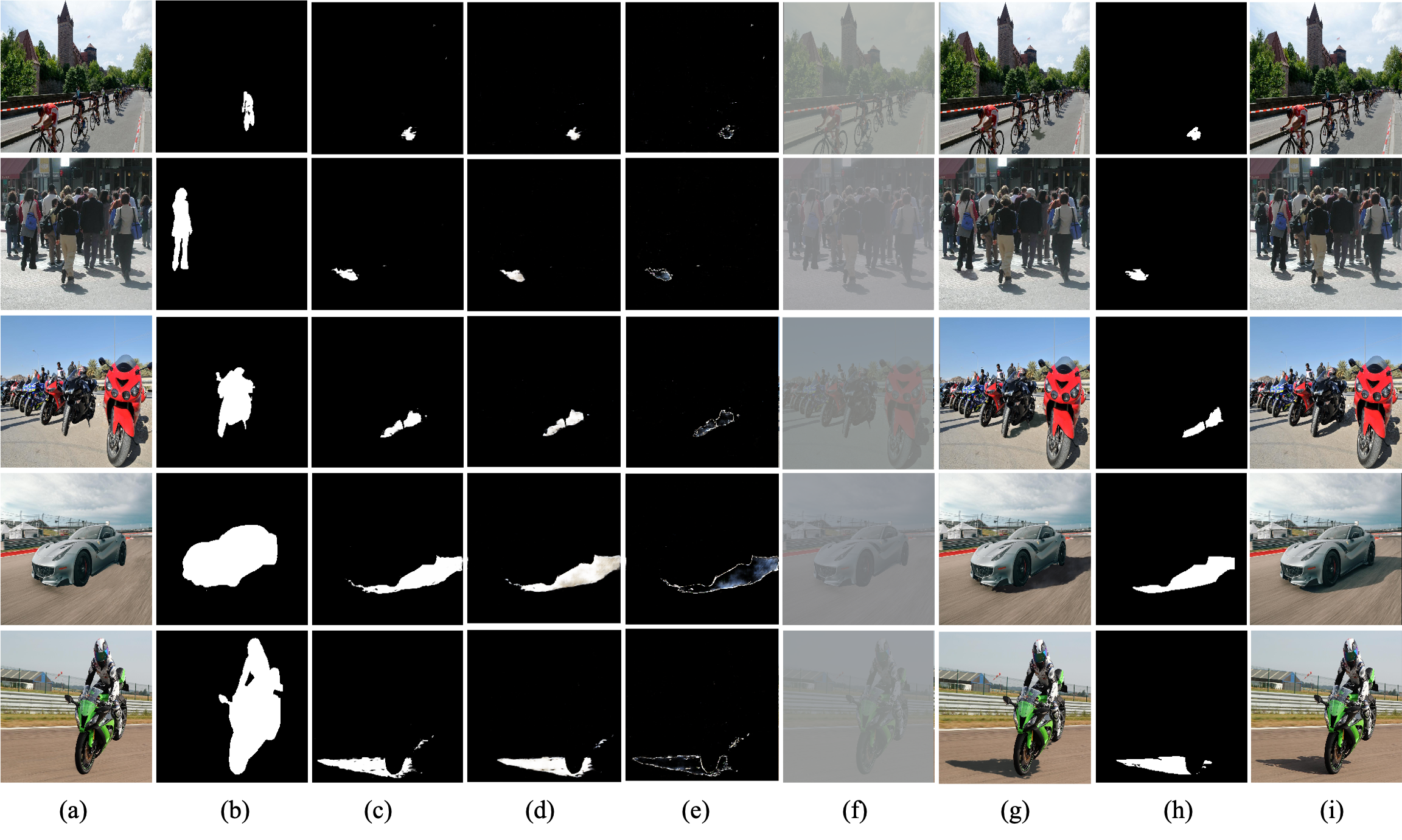}
\end{center}
\caption{Visualization of intermediate results produced by our SGRNet on DESOBA dataset. From left to right are input composite image $\bm{I}_c$ (a), foreground object mask $\bm{M}_{fo}$ (b), generated foreground shadow mask $\tilde{\bm{M}}_{fs}$ (c), generated shadow matte $\tilde{\bm{\alpha}}$ (d), the difference between generated foreground shadow mask $\tilde{\bm{M}}_{fs}$ and the generated shadow matte $\tilde{\bm{\alpha}}$ (e), generated darkened image $\tilde{\bm{I}}_c^{dark}$ (f), generated target image $\tilde{\bm{I}}_g$ (g), ground-truth foreground shadow mask $\bm{M}_{fs}$ (h), ground-truth target image $\bm{I}_g$ (i). The results on BOS (\emph{resp.,} BOS-free) test images are shown in row 1-3 (\emph{resp.,} row 4-5). }
\label{fig:intermediate} 
\end{figure*}

\section{The Impact of Shadow Size}\label{sec:immpact_shadow_size}
To investigate the impact of shadow size as well as the difference between BOS test images and BOS-free test images, we conduct experiments on DESOBA dataset by dividing test images into different subsets according to the foreground shadow ratio $r_{fs}$ (\emph{i.e.}, the area of ground-truth foreground shadow over the area of the whole image). 
In detail,  considering that the foreground shadow ratio is generally smaller than $0.08$, we firstly divide BOS and BOS-free test images into $2$ subsets: $r_{fs}>0.08$ and $r_{fs}<=0.08$. Then, the subset with $r_{fs}<=0.08$ is further divided into $3$ subsets: $r_{fs}\in(0,0.02]$, $r_{fs}\in(0.02, 0.04]$, and $r_{fs}\in(0.04, 0.08]$. In Table~\ref{tab:shadow_size}, we report the numbers of BOS and BOS-free test images in each subset and the results of BOS and BOS-free test images in each subset. It can be seen that GRMSE, GSSIM, and LSSIM generally become worse when $r_{fs}$ increases, because larger foreground shadow usually leads to worse results. 

When comparing the results on BOS-free test images and the results on BOS test images, we do not have consistent conclusions, because the difficulty of generating foreground shadow depends on many factors in a complicated and interlaced way. 
Although BOS test images have additional shadow clues provided by background object-shadow pairs, BOS-free test images usually have simpler background scenes and a single foreground object located at center, which may make the task easier. 

\section{Visualization of Ablation Studies}\label{sec:vis_ablated_methods}
We compare our full-fledged method with ablated versions:  ``w/o $E_{BS}$", ``w/o CAI", ``w/o $\bm{M}_{bos}$", and ``w/o Fill" (see Section $5.3$ in main paper). We visualize some  example results produced  by different ablated  methods on both BOS test imags and BOS-free images of DESOBA dataset in Figure~\ref{fig:abalation}. The comparison between these ablated methods and full method proves the advantage of fully exploiting background information and delicate network design based on illumination model. 


\begin{figure*}[t]
\begin{center}
\includegraphics[scale=0.69]{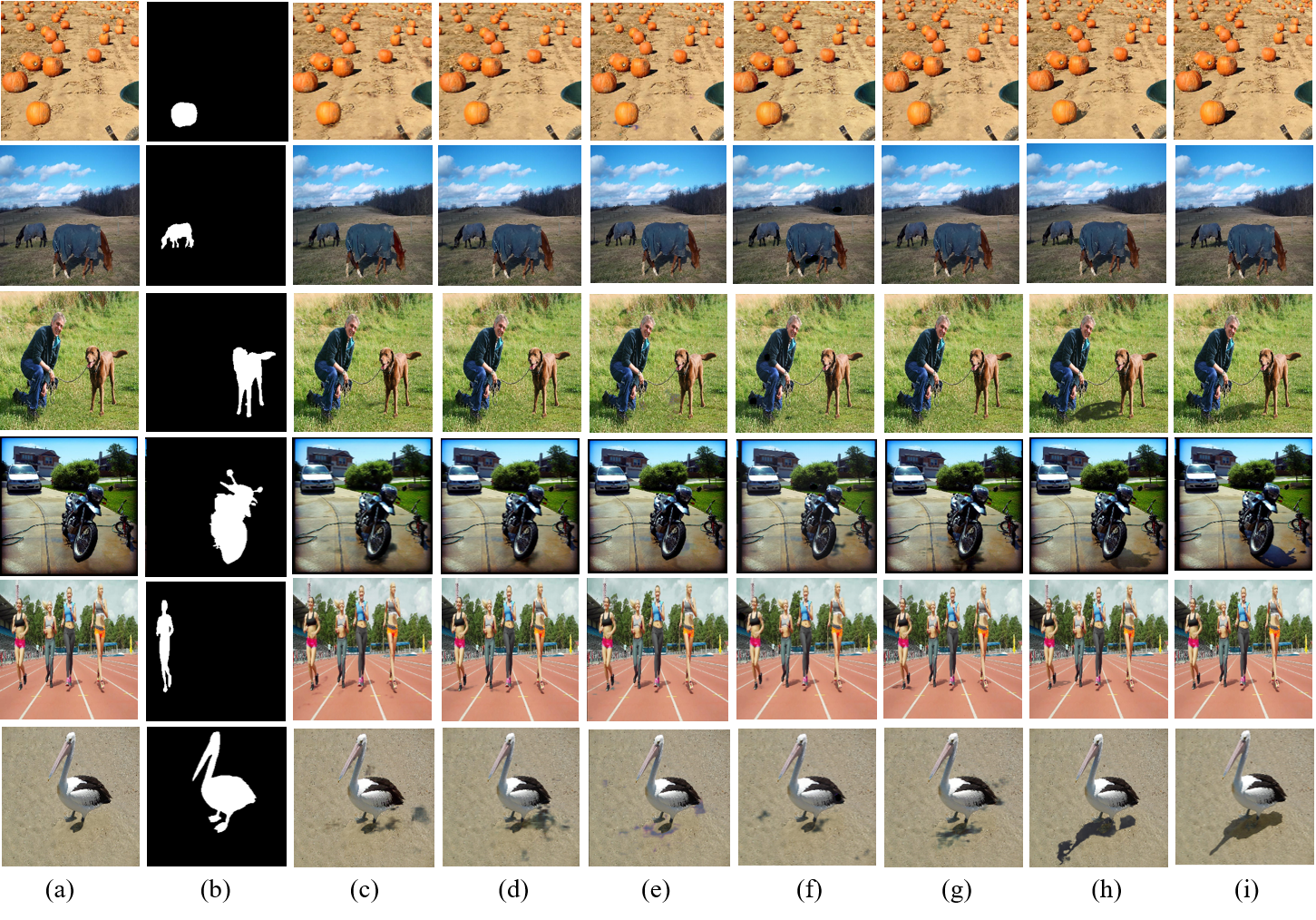}
\end{center}
\caption{Visualization comparison with different methods on DESOBA dataset. From left to right are input composite image (a), foreground object mask (b), results of Pix2Pix
(c), Pix2Pix-Res (d), ShadowGAN (e), Mask-ShadowGAN (f), ARShadowGAN (g), SGRNet (h), ground-truth (i). The results on BOS (\emph{resp.}, BOS-free) test images are shown in row 1-5 (\emph{resp.}, 6).}
\label{fig:desoba} 
\end{figure*}

\begin{figure*}[t]
\begin{center}
\includegraphics[scale=0.7]{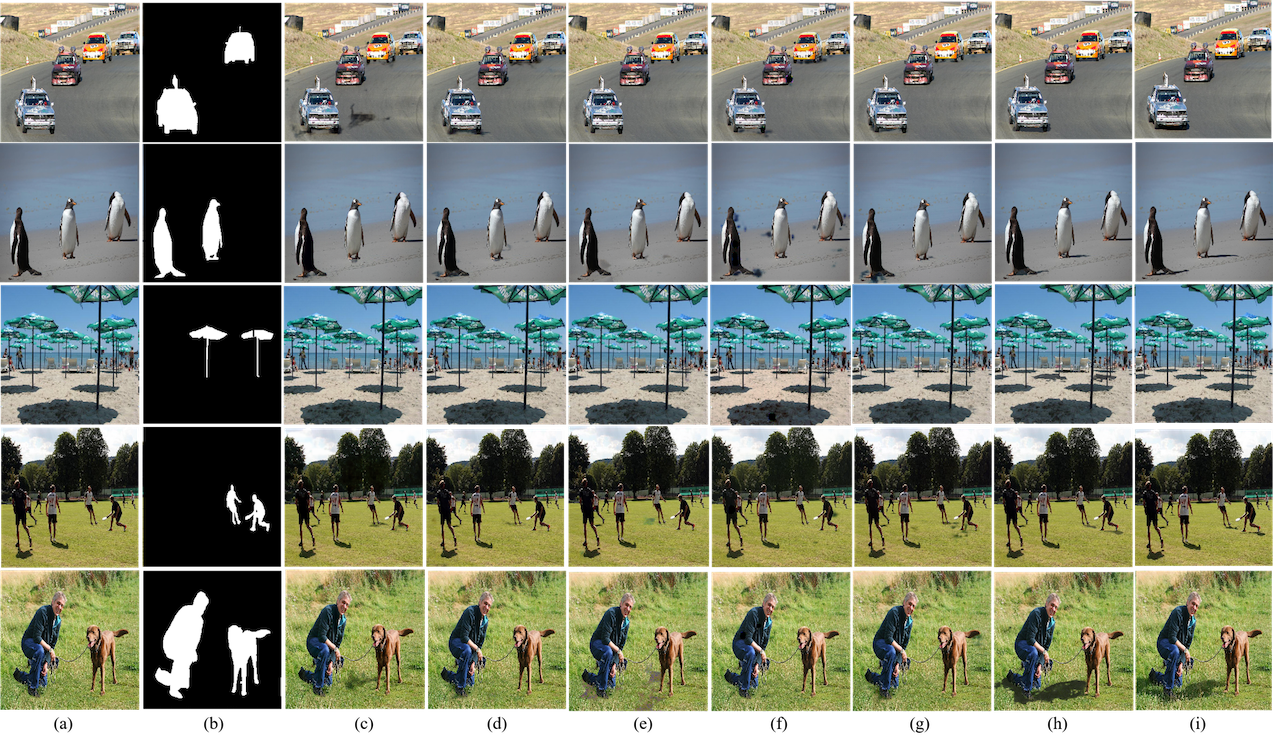}
\end{center}
\caption{Visualization comparison with different methods on synthetic composite images with two foreground objects from DESOBA dataset. From left to right are input composite image (a), foreground object mask (b), results of Pix2Pix
(c), Pix2Pix-Res (d), ShadowGAN (e), Mask-ShadowGAN (f), ARShadowGAN (g), SGRNet (h), ground-truth (i). The results on BOS (\emph{resp.}, BOS-free) test images are shown in row 1-4 (\emph{resp.}, 5).}
\label{fig:desoba_twoforeground} 
\end{figure*}

\section{Ablation Studies on BOS-free images}\label{sec:ablated_methods_bosfree}
We have provided the ablation studies on BOS test images from DESOBA dataset in Table 2 in main paper. Here, we additionally provide the ablation studies on BOS-free test images from DESOBA dataset. Same as Section 5.3 in the main paper, we analyze the impact of loss terms and alternative network designs of our SGRNet on BOS-free test images, and report quantitative results in Table~\ref{tab:ablation_bosfree}. Compared with ablative results on BOS images in main paper, we have roughly consistent observations on BOS-free test images, which demonstrates the effectiveness of each module in our method. 

\begin{figure}
\begin{center}
\includegraphics[scale=0.28]{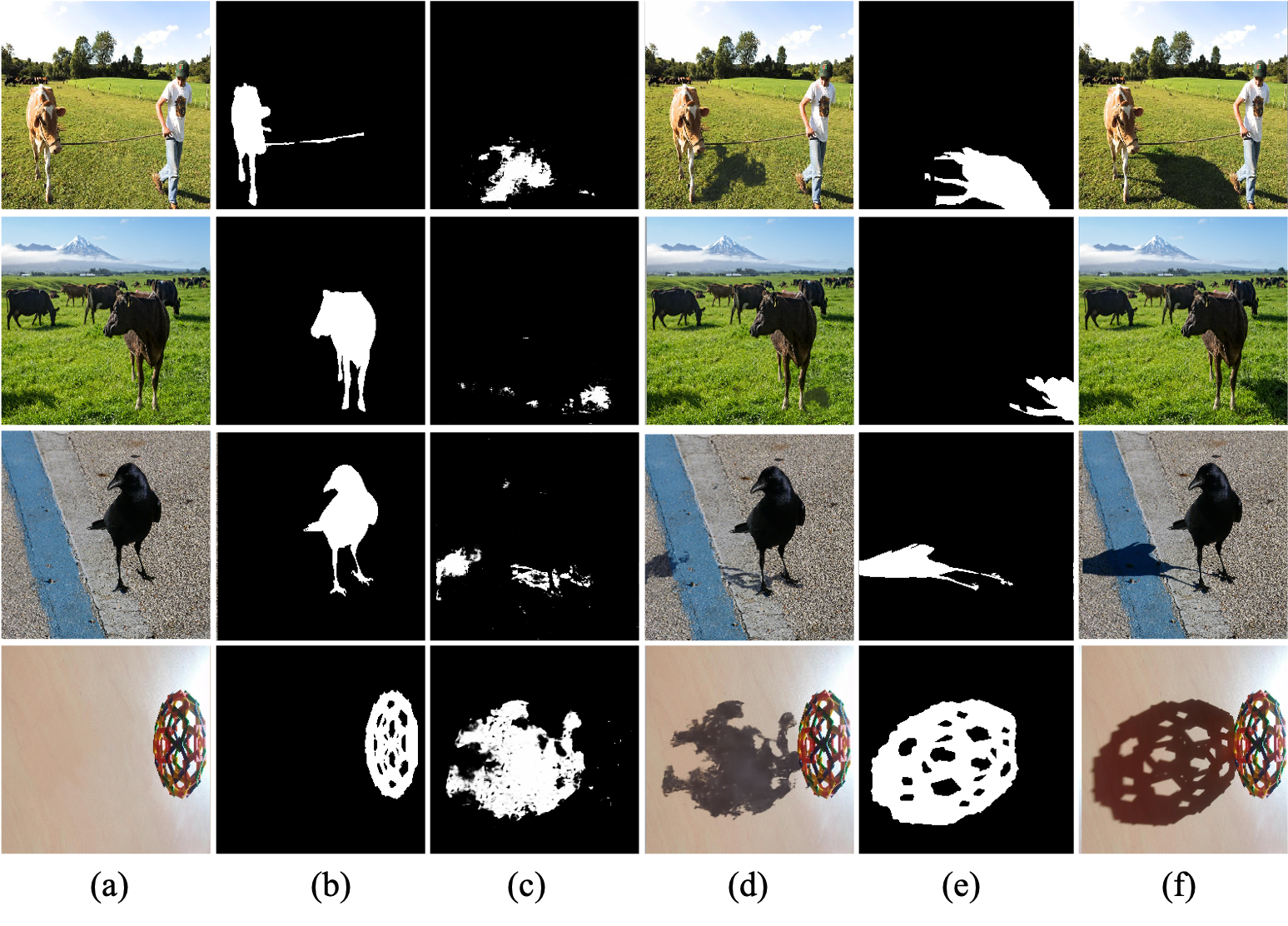}
\end{center}
\caption{Visualization of failure cases produced by our SGRNet. From left to right are input composite image $\bm{I}_c$ (a), foreground object mask $\bm{M}_{fo}$ (b), generated foreground shadow mask $\tilde{\bm{M}}_{fs}$ (c) , generated target image $\tilde{\bm{I}}_g$ (d), ground-truth foreground shadow mask $\bm{M}_{fs}$ (e), ground-truth target image $\bm{I}_g$ (f). The results on BOS (\emph{resp.}, BOS-free) test images are shown in row 1-2 (\emph{resp.,}  row 3-4).}
\label{fig:failure_cases} 
\end{figure}

\section{Visualization of Intermediate Results}\label{sec:vis_intermediate_results}
To demonstrate the effectiveness of our designed modules including shadow mask generator $G_S$, shadow parameter predictor $E_P$, and shadow matte generator $G_M$, we show some intermediate results. In Figure~\ref{fig:intermediate}, we show $\tilde{\bm{M}}_{fs}$ produced by $G_S$, $\tilde{\bm{I}}_c^{dark}$ produced after $E_P$, and $\tilde{\bm{\alpha}}$ produced by $G_M$. We can see that our shadow mask generator $G_S$ can infer the shape of foreground shadow. Our shadow parameter predictor $E_P$ can predict reasonable shadow parameters $\{\tilde{\bm{w}}^{dark}, \tilde{\bm{b}}^{dark}\}$, which are used to generate darkened images $\tilde{\bm{I}}_c^{dark}$. Our shadow matte generator $G_M$ can refine the foreground shadow mask and produce shadow matte of high quality.

\section{More Visualization Results on DESOBA Dataset} \label{sec:more_results_on_desoba}
We show more example images generated by our SGRNet and five baselines methods on DESOBA dataset in Figure~\ref{fig:desoba}. It can be seen that our SGRNet can generate shadows with plausible shapes, while other baseline methods even fail to produce any shadow. Moreover, our SGRNet can also produce foreground shadows with plausible shapes on BOS-free test images, which can be explained as follows. Even without paired object-shadow pairs in the background, our shadow mask generator can still attend relevant illumination cues (\emph{e.g.}, shading, sky appearance variation)~\cite{lalonde2012estimating, zhang2019all} from background image to predict the foreground shadow.


\section{Generated Results with Two Foreground Objects on DESOBA Dataset}\label{sec:two_foreground_object}
We show some example images with two foreground objects generated by our SGRNet and five baselines methods on DESOBA dataset in Figure~\ref{fig:desoba_twoforeground}. For synthetic composite images with two foreground objects, we can see that our SGRNet can generally generate foreground shadows with relatively reasonable shapes and shadow directions, while other baselines produce foreground shadows with implausible shapes, or none at all.

\section{Limitation}\label{sec:failure_case}
Since our DESOBA dataset is built on the real images, the background scenes and foreground objects could be very complicated. We show some failure cases in Figure~\ref{fig:failure_cases}. In row $1,2,3$, our SGRNet can only generate incomplete shadows. In row $4$, our SGRNet cannot capture the details of complex foreground object and thus fails to generate compatible foreground shadow.

\nolinenumbers
{
\bibliography{egbib.bib}
}